# VALIDITY ARGUMENTS FOR CONSTRUCTED RESPONSE SCORING USING GENERATIVE ARTIFICIAL INTELLIGENCE APPLICATIONS

**Jodi M. Casabianca, Daniel F. McCaffrey, Matthew S. Johnson, Naim Alper, and Vladimir Zubenko**

ETS

Corresponding author: jodi.casabianca@gmail.com

**January 4, 2024**

## ABSTRACT

The rapid advancements in large language models and generative artificial intelligence (AI) capabilities are making their broad application in the high-stakes testing context more likely. Use of generative AI in the scoring of constructed responses is particularly appealing because it reduces the effort required for handcrafting features in traditional AI scoring and might even outperform those methods. The purpose of this paper is to highlight the differences in the feature-based and generative AI applications in constructed response scoring systems and propose a set of best practices for the collection of validity evidence to support the use and interpretation of constructed response scores from scoring systems using generative AI. We compare the validity evidence needed in scoring systems using human ratings, feature-based natural language processing AI scoring engines, and generative AI. The evidence needed in the generative AI context is more extensive than in the feature-based NLP scoring context because of the lack of transparency and other concerns unique to generative AI such as consistency. Constructed response score data from standardized tests demonstrate the collection of validity evidence for different types of scoring systems and highlights the numerous complexities and considerations when making a validity argument for these scores. In addition, we discuss how the evaluation of AI scores might include a consideration of how a contributory scoring approach combining multiple AI scores (from different sources) will cover more of the construct in the absence of human ratings.

Author Note: This work was presented at the 2024 meeting of the International Testing Commission in Granada, Spain. We are grateful to Paul Jewsbury (ETS) for his careful review of this manuscript and thoughtful suggestions.

**Introduction**

Natural language processing (NLP) solutions for automatically scoring constructed responses (CR) are well established and used broadly in standardized testing for written, spoken, and short answer responses. In many applications, a set of features are selected that are intended to represent the construct as defined by the scoring rubric and combined to predict human ratings. These features are handcrafted and trained by NLP scientists to be extracted from the response and then used in a model to generate an overall score for the response. Some new artificial intelligence (AI) solutions, specifically approaches using generative AI such as GPT4, are not engineered to produce features based on the same principles of NLP to match the scoring rubric and construct. Instead, generative AI approaches involve prompting an underlying large language model (LLM) to produce outputs such as a rating, sometimes with relatively little training or fine-tuning of the LLM to the specific task of scoring. It is difficult to explain how these generative AI approaches obtained their outputs since the LLM has billions of parameters, but they do offer capabilities that were not previously available without extensive effort by experts.

While the future of assessment will undoubtedly make extensive use of AI and given the accessibility of LLMs, it is important to place guardrails around their use in educational assessment so that they are used responsibly. Discussions on ethical AI and standards for using AI in education are prevalent in the literature (Bulut et al., 2024; International Test Commission & Association of Test Publishers, 2022; Johnson, 2024). This paper describes a study exploring the utility of GPT4 in the context of CR scoring to highlight how validity arguments might be specially structured when using generative AI. The following sections summarize generative AI and discuss established validity frameworks for evaluating the human ratings and automated scores from traditional CR scoring systems. We then introduce an additional set of validity evidence that should be collected and documented when using generative AI in CR scoring systems. We demonstrate the curation of validity evidence using empirical data from three standardized tests historically scored with humans and automated scoring engines. We conclude with suggestions for practitioners.

**The Basics of Generative AI**

The traditional use of AI in CR scoring relied on experts such as NLP scientists and linguists to create hand-crafted features quantifying different components of written text or spoken responses (Shermis & Burstein, 2013). The NLP scientists' expertise in creating and purposefully combining features to provide construct coverage and alignment with the scoring rubric ensured a human-in-the-loop approach to AI scoring. The statistical models once used to combine features were simple, for instance, multiple linear regression.

This traditional use is in stark contrast to generative AI, which is a form of AI in which a LLM or another type of model (e.g., generative adversarial networks [GANs], recurrent neural networks [RNNs]) produces or generates responses to *prompts*. For example, a prompt could request the LLM provide a score to written responses according to a rubric yielding an AI scoring system with minimal human inputs. The LLMs are based on statistical patterns extracted from extremely large datasets which might be produced from a combination of NLP corpora, including annotated texts, and extracts of text from the internet. The most popular LLMs use millions or billions of parameters, which is why some label LLMs as "black-box" algorithms. Compared to smaller language models, LLMs can typically perform more tasks with better results because the very large number of parameters and training on massive data give the model exceptional flexibility to solve even complex tasks. Often the end user of the LLM knows little about the training data or the inner workings of the LLM at all because they did not build the LLM. While the end user may use the LLM to score essays or spoken responses, they might not know how or why it produces the scores it does.

There exist different scoring approaches in between the extremes of using handcrafted features linked to a construct definition to predict human ratings and prompting an LLM. Figure 1 places these different scoring approaches on a continuum to demonstrate the reduction in transparency. To predict a human rating, we might use machine learning methods to extract general linguistic features and/or keyword indicators to capture specific content in the response or specific written structures not explicitly related to the content. This approach is often used in scoring short answer texts or content-based items. We might also use embeddings from an LLM to predict the human ratings. The transparency in these types of models is reduced compared to the substantive, construct-related feature-based model, but better than when prompting an LLM. What distinguishes these modeling approaches to prompting LLMs is that they are intended to predict the human rating, and there is an expectation of some concordance with human scoring behavior. Unless the LLM is fine-tuned for used human ratings, there is no explicit connection to human ratings.

To set up some language to further the discussion of LLMs, let the information we give to the LLM be called "prompts" which contain instructions for the request of the LLM and any information needed to complete the "task" or the request. An example of a prompt in the CR scoring context is given in the Appendix of this paper. The result is often referred to as the "completion", but we will often refer to it as the output, result or the LLM score, as appropriate.

The LLM output also depends on various technical options that can influence the results. Some such options are built into an LLM (e.g., the algorithm for tokenizing the textual inputs or the algorithm for selecting words for a completion) but can vary across models leading them to be differentially effective for a given application. Others, such as setting the temperature, which affects the variability of potential responses, can be controlled by the users. Users need to explore these settings to ensure the LLMs yield the most reliable scores and ones that can best be shown to adhere to the construct when using LLMs for scoring responses.

**Figure 1**
*Approaches to Scoring Constructed Responses in Order of Reduced Transparency*

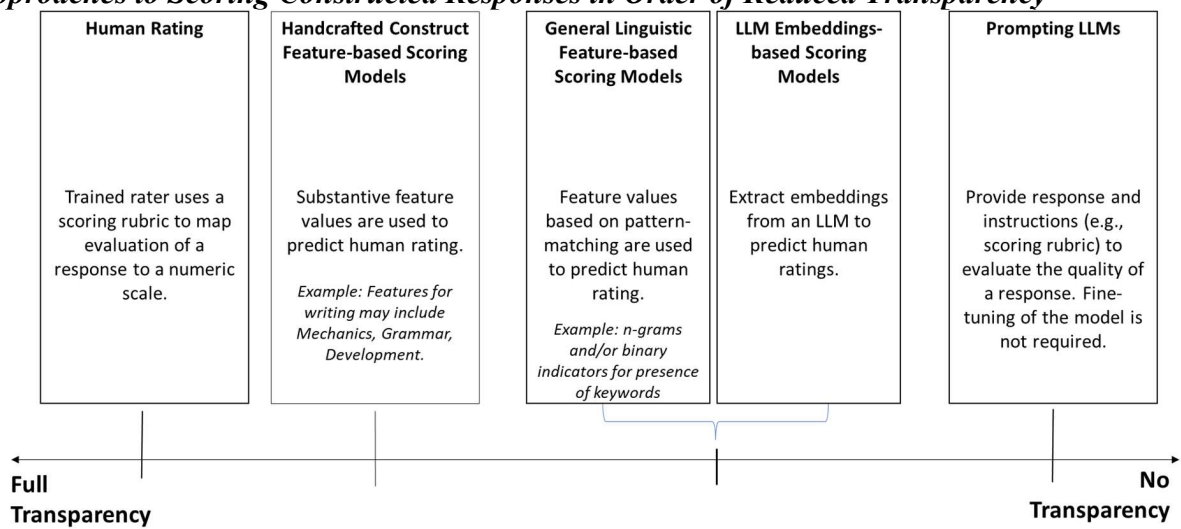

**Training LLMs**

LLMs are most often pre-trained by the model providers (e.g., OpenAI, Meta, etc.). Post-training, which can include fine-tuning and retrieval augmented generation (RAG), often helps to align a model closer with a specific use case, which could make it more appropriate for a particular task or domain. In the process of fine-tuning, the user can provide examples of responses with the "correct" scores or labels (prompt-completion pairs), to train the LLM on how to perform a specific task, updating the weights in the LLM. This is similar to model building in the traditional sense in that we might use a sample of responses and the human ratings to "train" the "engine" to estimate model weights for future prediction. The difference here is that the LLM could be used to produce scores without fine tuning for the specific use case.

While the process of fine-tuning seems like it would be easy to perform and could only improve the model, there are certain considerations. For example, if the model is fine-tuned to perform task "A" and is then later used to perform task "B" performance might be degraded compared to the original pre-trained (not fine-tuned) model. The user must be aware of both the limitations of the fine-tuned model and of existing alternatives if the fine-tuned model will be used for multiple different tasks, such as scoring different item types. Parameter efficient fine-tuning (PEFT) trains a small number of specific "adapter" layers of the LLM. In the case of PEFT, most of the pre-trained weights remain the same and so it is robust for performing multiple tasks and prevents overfitting. Fine-tuned LLMs are readily available for certain purposes and tasks. To perform additional domain-specific fine-tuning in the CR scoring context, we might use response data scored or annotated by humans collected as part of testing operations, and/or publicly-available datasets.

Another way to train the model would be in-context learning (ICL) which occurs during the prompting. For example, we might provide the LLM with one or more prompt-completion pairs as examples of what it is supposed to do with the subsequent prompts for which we want a completion. Since there is a limit to what can be submitted to the LLM in this fashion, typically only up to three examples may be provided. If no examples are used, we call this zero-shot prompting, if one example, then it is termed one-shot learning, and if two or three, few-shot learning.

**Evaluation for Accuracy and Fairness**

The high expectations for the functionality of LLMs, concerns about potential risks or harms from their use, and the existence of many alternative models has led to evaluation of the actual capabilities of LLM as an active research area. Evaluation often relies on how well the models perform on sets of benchmark tasks in terms of the accuracy, repeatability, and fairness of the results on each task. Because of the broad range of potential uses for LLMs, some authors argue for moving testing performance on specific tasks to using a "psychometric" approach for LLM evaluations that assesses the models on underlying types of tasks that might be analogous to tasks in psychology such as spatial reasoning tasks (Bommasani at al., 2023). For example, in medical and health use cases, ratings scales are used by human raters to evaluate LLMs on dimensions such as accuracy, understanding, safety, and trust (Tam et al., 2024)

The evaluation of LLMs for rating constructed responses has not received specific attention in the research on evaluation of LLMs. However, when used for predicting human ratings as a means of scoring CRs, LLMs have been evaluated using the approaches developed for more traditional feature-based AI scoring that uses simpler statistical and machine-learning models. These evaluations generally focus on how well the models recover the human ratings and the ability to do this consistently regardless of the characteristics of the test taker who produced the response. The evaluations use common statistics based on agreement (e.g., simple percent agreement between the scores of humans and AI models, and kappa and quadratic weighted kappa [QWK]) and accuracy (e.g., mean squared error [MSE] and percent reduction in mean squared error [PRMSE]). Some methods, notably PRMSE, account for the errors in the human ratings used in assessing the AI models. Evaluations also include checks for fairness (Johnson & McCaffrey, 2023; Johnson et al., 2022). In the psychometrics and measurement community, the evaluation of the quality of the model prediction is embedded in the larger framework of demonstrating the validity of the scores for supporting the claims of the items and tests (Bennett & Zhang, 2015; McCaffrey et al., 2022; Williamson et al., 2012).

Beyond metrics, explainability tools help evaluate LLMs by revealing what parts of the prompt/input are important in the resulting completion (or score). In the black-box algorithms, there is no possibility of purposefully connecting the inner workings of the LLM and the score it assigns like we can in a feature-based model. AI explainability (XAI) and explainable NLP (XNLP; Danilevsky et al., 2020) are approaches to generating explanations for the LLM results. Recent research has led to tools that can be used to identify features inside an LLM to explain how it works and classify those features as "safety relevant" or harmful. They may be generated in different ways including, for example, use of measures of feature importance, surrogate modeling, or example-driven approaches, and the results may be visualized in different ways. A popular method is using saliency maps or highlighting to show which tokens in the text are important to prediction accuracy and visualize the gradient of the loss with respect to each token in the model. The maps provide indicators of the influence of the tokens in the response. Given the available tools, the explanations are not always useful and come with their own scrutiny and need for evaluation (Hoffman et al., 2018).

XAI tools can be used to evaluate fairness in addition to the traditional metrics that might be used such as subgroup-level comparisons of human and machine scores. The evaluation of fairness is very important when using pre-trained LLMs as they are typically trained using data not representative of the existing cross-cultural psychological diversity of many tested populations (Atari et al., 2024). De-biasing models may be performed by some developers of LLMs, but the extent to which this is effective in any specific application is variable.

## Validity Framework for CR Scoring

The Standards for Educational and Psychological Testing (American Educational Research Association [AERA], American Psychological Association [APA], & National Council on Measurement in Education [NCME], 2014) provide high level guidelines for demonstrating the validity (including fairness) of CR scoring. Even though the document was written when the use of NLP and AI for scoring was just emerging into the mainstream, the Standards addressed automated scoring on three occasions. Its guidelines can serve as the basis for evaluating validity for scoring constructed responses with generative AI. Other publications such as Bennett and Zhang (2015), McCaffrey et al. (2022), and Williamson et al. (2012) elaborated approaches for collecting validity evidence for constructed response scores from human raters and feature-based AI scores. We build on these to create a framework for generative AI scoring, first reviewing the frameworks for human ratings and feature-based AI scoring and then extending those guidelines to generative AI.

**Evidence for Human Ratings**

To provide more detailed guidance on collecting evidence for AI scores, ETS published its *Best Practices for Constructed-Response Scoring* (McCaffrey et al., 2022) which included a framework for establishing validity evidence for CR scores from human raters or AI. The validity evidence for CR scores, of any kind, starts with the task design. Ideally task and test design are conducted within a formal framework such as evidence-centered design (ECD; Mislevy et al., 2003). Part of task design for CR items is the scoring rubric for judging the responses and assigning scores. When humans provide the scores, the key to the validity of the scores is the ability of the human raters to consistently use the rubric as intended. To ensure this occurs, there are several processes involved with managing human rating. Decisions made throughout the design of the scoring system should be made with principles of validity, reliability, and fairness, and they should be documented as part of the curation of validity evidence. For example, the training materials developed for raters should include additional details on the task including annotated exemplars to ensure that raters will apply the rubric as it was intended. Raters should be trained and then evaluated before scoring to ensure they are qualified to score. To the extent possible, raters should be recruited from a diverse population with a specified skill set qualifying them to rate responses for the assessment. During the rating process, raters should be monitored to make sure that they are consistent and accurate. This means that during the design of the system before scoring occurs, decisions must be made on how to collect data to measure consistency and accuracy. Example of such decisions are: What size sample of double-scored responses is necessary to estimate the interrater agreement using the selected statistic? Will exemplar responses (pre-selected with an agreed upon score from experts) be used to track rater accuracy, and how will they be selected and interwoven into operational scoring? These design decisions should be made in a principled fashion to ensure that all components of the human rating process contribute to meaningful use and interpretation of test scores.

Table 1 provides a list of five different types of evidence that should be collected to make a validity argument for scores (as per the 2014 *Standards*) and examples of specific pieces of evidence that should be documented and/or collected to make the argument for scores based on human ratings. For example, evidence of internal structure might include a documented link between the item or task and the construct definition. For a CR item, there should also be a documented link between the scoring rubric and the construct definition. This documentation may include a memorandum or report summarizing how the task allows the test taker to demonstrate knowledge, skills and abilities related to certain aspects of the construct. Other traditional evidence, such as factor analyses, are included here. Importantly, we should demonstrate that the human rating process minimizes construct-irrelevant variance and therefore a system for monitoring the interrater agreement and accuracy should be planned, implemented, and documented.

Other types of evidence have been heavily discussed in the literature and so have the methods used to detect unfairness, which are included in the last set of rows of Table 1 as a type of evidence to create a validity argument. For example, validity coefficients (e.g., correlations) from predictive validity studies (as appropriate), inter-item correlations, and content reviews are standard validity evidence collected for selected responses and CR items and their approaches are well-established. However, the human rating component introduces added opportunities for

**Table 1**
*Validity Evidence for CR Scoring Systems*

| Type of Validity Evidence | Human Ratings | Construct Feature-based AI Scores | General Linguistic Features-based and Embeddings-based AI Scores | Generative AI Scores |
|---|---|---|---|---|
| **Internal Structure** | Link between prompt and construct definition | | | |
| | Link between scoring rubric and construct definition | Established connection between features and rubric (and construct definition) | | |
| | Training materials, exemplars, calibration test, certification process, etc. designed to be aligned with construct (and have no construct irrelevant features) | Features trained on representative sample not used for model-building | If applicable, features trained on representative sample not used for model-building. | Documentation on prompting strategy, in-context learning, and fine-tuning decisions. |
| | Factor analysis to confirm internal structure of test and/or correlation analysis (inter-item correlations) | Factor analysis to confirm internal structure of test based on engine scores and/or correlation analysis (inter-item correlations) | Factor analysis to confirm internal structure of test based on engine scores and/or correlation analysis (inter-item correlations) | Factor analysis to confirm internal structure of test based on engine scores and/or correlation analysis (inter-item correlations) |
| | Document the support for the selection of the raters (expertise, experience, etc.) | Document the support for the selection of the scoring engine/feature set | Document the support for the selection of the scoring engine/feature set and/or LLM embeddings | Document the support for the selection of the LLM |
| | Evaluation of ratings for rater accuracy and interrater consistency | Concordance between human ratings and machine scores (initially during model evaluation and ongoing monitoring) | Concordance between human ratings and machine scores (initially during model evaluation and ongoing monitoring) | Concordance between human ratings and machine scores (initially and ongoing monitoring), given a |

| Type of Validity Evidence | Human Ratings | Construct Feature-based AI Scores | General Linguistic Features-based and Embeddings-based AI Scores | Generative AI Scores |
|---|---|---|---|---|
| | | | | sufficient sample of human ratings.<br><br>Studies showing reproducibility and consistency of LLM scores over time. Documentation of the variability of LLM scores and how that affects reported score reliability. |
| | | Expert review of scores and responses at all score levels | Expert review of scores and responses at all score levels | Expert review of scores and responses at all score levels |
| | | | | Analysis of chain-of-thought output from LLM to consistency with construct definition. |
| **Relations to External Variables** | Moderate correlations with section and total scores | Moderate correlations between machine scores and section/total scores; comparison to evidence based on human ratings | Moderate correlations between machine scores and section/total scores; comparison to evidence based on human ratings | Moderate correlations between machine scores and section/total scores; comparison to evidence based on human ratings |
| | Moderate correlations with other tests and external variables | Moderate correlations between machine scores and other tests and external variables; comparison to evidence based on human ratings | Moderate correlations between machine scores and other tests and external variables; comparison to evidence based on human ratings | Moderate correlations between machine scores and other tests and external variables; comparison to evidence based on human ratings |

| Type of Validity Evidence | Human Ratings | Construct Feature-based AI Scores | General Linguistic Features-based and Embeddings-based AI Scores | Generative AI Scores |
|---|---|---|---|---|
| | Convergent/discriminant validity studies / validity coefficients | Convergent/discriminant validity studies / validity coefficients based on machine scores | Convergent/discriminant validity studies / validity coefficients based on machine scores | Convergent/discriminant validity studies / validity coefficients based on machine scores |
| | If applicable: Contrasted group studies, predictive/concurrent validity studies / validity coefficients | If applicable: Contrasted group studies, predictive/concurrent validity studies / validity coefficients | If applicable: Contrasted group studies, predictive/concurrent validity studies / validity coefficients | If applicable: Contrasted group studies, predictive/concurrent validity studies / validity coefficients |
| **Response Processes** | Review of response processes for the item | | | |
| | Review of response processes for the rubric (e.g., thinkalouds for raters) | Expert annotations of responses scored by machine at each score level | Expert annotations of responses scored by machine at each score level | Expert annotations of responses scored by machine at each score level |
| | | Expert review of atypical responses and evaluation of AI scores | Expert review of atypical responses and evaluation of AI scores | Expert review of atypical responses and evaluation of AI scores |
| | | | | Expert evaluation of chain of thought feedback and comparison to expert annotation, if available. |
| **Test Content** | Expert review of prompt and rubric to demonstrate content coverage and no construct irrelevance | Expert review of features and weights to demonstrate content coverage and that there are no features that weigh too heavily or are misaligned with construct/rubric | | |

| Type of Validity Evidence | Human Ratings | Construct Feature-based AI Scores | General Linguistic Features-based and Embeddings-based AI Scores | Generative AI Scores |
|---|---|---|---|---|
| | Inter-item correlations and correlations between item and section scores | Inter-item correlations and correlations between item and section scores. Comparison of these correlations based on human ratings and machine scores. | Inter-item correlations and correlations between item and section scores. Comparison of these correlations based on human ratings and machine scores. | Inter-item correlations and correlations between item and section scores. Comparison of these correlations based on human ratings and machine scores. |
| | | Expert annotations of responses scored by machine at each score level | Expert annotations of responses scored by machine at each score level | Expert annotations of responses scored by machine at each score level |
| Consequence of Use | Analysis of unintended consequences of test score use including test taker morale, access to education, academic/career pressure and anxiety. Collected via surveys or other studies of test takers, instructors, etc. | Analysis of unintended consequences of test score use including test taker morale, access to education, academic/career pressure and anxiety. Collected via surveys or other studies of test takers, instructors, etc. A comparison of unintended consequences based on human scores and AI scores. | Analysis of unintended consequences of test score use including test taker morale, access to education, academic/career pressure and anxiety. Collected via surveys or other studies of test takers, instructors, etc. A comparison of unintended consequences based on human scores and AI scores. | Analysis of unintended consequences of test score use including test taker morale, access to education, academic/career pressure and anxiety. Collected via surveys or other studies of test takers, instructors, etc. A comparison of unintended consequences based on human scores and AI scores. |
| | Analysis of intended consequences as a result of test score use including instructional changes, reclassifications, admissions, etc. | Analysis of intended consequences as a result of test score use including instructional changes, reclassifications, admissions, etc. A comparison of | Analysis of intended consequences as a result of test score use including instructional changes, reclassifications, admissions, etc. A | Analysis of intended consequences as a result of test score use including instructional changes, reclassifications, admissions, etc. A comparison of |

| Type of Validity Evidence | Human Ratings | Construct Feature-based AI Scores | General Linguistic Features-based and Embeddings-based AI Scores | Generative AI Scores |
|---|---|---|---|---|
| | | intended consequences based on human scores and AI scores. | comparison of intended consequences based on human scores and AI scores. | unintended consequences based on human scores and AI scores. |
| **Fairness** | Item fairness reviews | | | |
| | Differential item functioning analysis | Differential item functioning analysis | Differential item functioning analysis<br><br>Consult with subject matter expert on saliency results to understand if the scores are biased. | Consult with subject matter expert on saliency results to understand if the scores are biased. |
| | Evaluate the fairness of human rating quality by subgroup (human-human agreement by subgroup compared to overall) | Evaluate the fairness of human ratings used to train the model | Evaluate the fairness of human ratings used to train the model | Evaluate the fairness of human ratings used to fine-tune the model (if applicable). |
| | | Collect a subset of human ratings and perform traditional statistical comparison (e.g., human-machine SMD, etc.). | Collect a subset of human ratings and perform traditional statistical comparison (e.g., human-machine SMD, etc.). | Collect a subset of human ratings and perform traditional statistical comparison (e.g., human-machine SMD, etc.). |
| | | Differential feature functioning analysis | Use saliency methods to understand differences in responses and assigned scores for different subgroups. | Use saliency methods to understand differences in responses and assigned scores for different subgroups. |

| Type of Validity Evidence | Human Ratings | Construct Feature-based AI Scores | General Linguistic Features-based and Embeddings-based AI Scores | Generative AI Scores |
|---|---|---|---|---|
| | | Differential algorithmic bias analysis | Differential algorithmic bias analysis | Differential algorithmic bias analysis |
| | | | | Report on established fairness metrics for the LLM. |
| | | | Report on the data used to pre-train the LLM. | Report on the data used to pre-train the LLM. |

construct irrelevant variance. In CR scores, for example, we want to expand the traditional sense of response processes by adding the response process of the rater using the rubric.

## Evidence for Automated Scores from Models Predicting Human Ratings

### *Construct Feature-based Models*

For construct feature-based AI scores, there is a chain of evidence that goes from the automated scores back to the human ratings due to how the engine is trained. Thus, it is important to make a validity argument for the human ratings and then the machine scores. Ideally, tasks that are intended to be automatically scored are designed with automated scoring in mind, and the procedures and decision-making should also be documented. If human ratings are then used to train automated scoring models, we prefer to start with "high quality" ratings based on sound practices as described above and in Table 1 and with at least a satisfactory psychometric profile.

The traditional NLP feature-based automated scoring approach requires similar evidence to the human ratings with some differences due to the nature of the scoring process. While human raters apply a scoring rubric and make judgements, a scoring engine extracts information from the responses to reflect different construct-relevant features. A statistical prediction model is then trained to predict the human rating using those features. In an engine that evaluates writing ability, features may include grammar, usage, mechanics, style, and organization (Attali, 2007; Attali & Burstein, 2006). In an engine that evaluates spoken responses, features might include words per minute, average pause length and others for accuracy and pronunciation (Xi et al., 2008). The set of features should be combined to represent the construct. Validity evidence may include documented links between the feature set and the construct definition. It may also include a summary of the prediction model weights to determine the extent to which the combination of features and weights correspond to how raters should be combining information about the response to derive the score.

For evidence of internal structure of the test (AERA, APA, & NCME, 2014, p. 16), we would collect evidence analogous to what we would collect for the human ratings—documented links between the features and scoring rubric; we might perform factor analyses or inter-item correlations with the machine scores and other item scores on the test; and we also monitor the AI scores by comparing them to human ratings of the same responses. This concordance should be established during model evaluation, but also during operational scoring to ensure there are no issues with predictions on a new sample of test takers. In addition, we should train the features on data not used for model-building or evaluation and the sample should be representative of the target test taker population.

Much of the other types of evidence for AI scores would be collected at the time of the initial model evaluation. A model-level evaluation includes an examination of concordance between the human ratings and the machine scores. Metrics might include standardized mean differences, correlations, QWK, etc. (Williamson et al., 2012). Recent arguments have been made to include a measure of the accuracy of predicting the human true score via PRMSE (McCaffrey et al., 2024).

An impact-level evaluation might compare the CR section scores (based on machine scores) to other section scores, or other completely external scores if available. We also want to compare section and total test scores based on the machine scores to corresponding section and total scores based on human ratings to understand the size of the differences at that level. For evidence that the response processes are appropriate, we rely on annotations by subject matter experts of a selection of responses to make sure that there is justification for the scores given by the engine. This type of qualitative analysis should be performed at all score levels. In addition, we also must provide evidence that the engine is properly handling "atypical" responses which may be in the wrong language, off-topic, a copy of the prompt text, etc. The scoring engine must be trained to either flag these responses so that they can be hand-scored or trained to assign an appropriate score to them (likely a 0). Analyzing the effectiveness of any "advisories" or flags the engine might use is important to ensure that these responses are detected and scored appropriately. This minimizes the chances that atypical responses that deserve a score of 0, for example, do not get a score of 2.

Checking for fairness is of vital importance, however simply checking for fairness is insufficient. Fairness should be part of the *design* of the scoring system. To start, in addition to the fairness checks that should occur on the human ratings, we need to ensure that the samples used to train features, build the models, and evaluate the models are large and representative of the test taker population. This is important because there might be differential response styles by group and the engine must be trained to score those appropriately. In addition, if there is an imbalance in the composition of the training sample and many groups constitute only a small proportion, there may be inadequate representation. Oversampling or weighting up small groups in the training sample may be a solution to this. An evaluation of the demographic composition of the sample and how it might impact fairness should be documented.

Fairness checks during model evaluation often involve a comparison of human and machine score means, by subgroup, via a standardized mean difference (Williamson et al, 2012). In addition, comparing the QWK by subgroup may also be helpful to understand if agreement is degraded for different test taker groups. The *Best Practices* (McCaffrey et al., 2022) discuss challenges with subgroup analyses, including that the small sample sizes for groups would prevent these groups from being assessed for fairness. Recent work utilizes empirical Bayesian methods to better estimate SMD in the small sample case (Kwon et al., n.d.). In addition to basic SMDs and QWKs, we might also run differential item functioning (DIF) analyses—first for the human scores and then for the machine scores, to compare the DIF results for humans and machines. Differential feature functioning (Zhang et al., 2017) and differential algorithmic functioning (Suk & Han, 2024) are also appropriate analyses to better understand the extent to which the engine features or the model may be disadvantaging certain groups. We may perform all of these together to collect evidence on the fairness of scores.

### Models Using General Linguistic Features and LLM Embeddings

Figure 1 shows three different types of models that predict the human rating: (a) construct feature-based model, (b) general linguistic feature-based model and (c) a model based on LLM embeddings. We include (b) and (c) in this section with (a) despite their relative reduced transparency because they share the chain of evidence link of the prediction of human ratings. This somewhat mitigates the concern with reduced transparency, even though there might still be many thousands of parameters and an increased risk of the predictions relying on construct irrelevant features of the response.

Most of the evidence required for these models will be similar to the construct feature-based models but not all will be available. For example, it may be infeasible or inappropriate to match engine features to rubric indicators. In content engines such as ETS' c-rater (Leacock & Chodorow, 2003), often a written response is evaluated using many generic linguistic features and/or keyword indicators to capture specific content in the response or specific written structures not explicitly related to the content. The responses may be parsed into n-grams which could generate thousands of "features". In this case, the features are not necessarily understandable, and the machine learning models used for prediction are more sophisticated than traditional regression models meaning the "weights" are not something that can be easily examined or immediately understood. The same would be true if using embeddings from an LLM. As a result, the transparency of the automated scoring process is reduced and the validity evidence for the machine scores relies more heavily on the quality of the human ratings used to train the models (McCaffrey et al., 2022). We might also require a very strong link between the scoring rubric and the construct definition and demonstrated high agreement between the human rating and AI scores during model evaluation and monitoring.

Some of the evidence we propose for generative AI scores will be useful for making a validity argument for these scores as well. Table 1 reflects these differences. For example, saliency methods to understand the importance of different features or embeddings may provide evidence that the scores resulting from these models lead to meaningful interpretations or that they do not contain construct irrelevant features.

## Validity Evidence for CR Scoring Systems Using Generative AI

Generative AI is distinctly different due to the nature of the approach in generating the scores. In this case, because the "engine" is generating an output in response to a prompt, there is no principled or explicit system of deriving a model or selecting features and the score is not based on a prediction of a human rating. As such the types of validity evidence we might collect for feature-based AI scoring should be expanded for generative AI applications.

### Choice of LLM

Much like any automated scoring engine, we should consider the goals of the assessment and the construct definition as we select a LLM to use for scoring. A written justification for the LLM is part of the evidence we should collect. A LLM is pretrained based on corpora and scrapes of text from the internet. Some LLMs are pretrained for specific domains, like domain-specific language (e.g., legal, medical, or some other content specific language). We might ask, is this LLM suited to the language task needed for the construct or use case? For example, a testing company evaluating responses for doctors or doctors-in-training might benefit from a domain-specific LLM, or from training their own. In addition, some LLMs are meant to generate text, not evaluate an input. If the task is to generate feedback on essay responses (e.g., annotations for training or for raters to use during scoring), then a LLM that was trained to generate text, such as GPT4, might be a suitable choice. If the task is to evaluate the text provided in the prompt, then other LLMs may be more appropriate. For example, GPT and BLOOM models are pretrained for text generation and other "emergent behavior" (Le Scao et al., 2023; OpenAI, 2023) while BERT and ROBERTA are pretrained for sentiment analysis, word classification, and named entity recognition (Devlin et al., 2018; Liu et al., 2019). T5 and BART are pretrained for translation, test summarization and question answering (Lewis et al., 2019; Raffel et al., 2020). As more LLMs are released for public use, it is important to have a good understanding from the user perspective on how they compare in their capabilities to fulfill the required task.

Selection of LLM should be based on empirical findings from preliminary experiments with multiple LLMs. Combinations of LLMs in AI scoring may also be considered. For example, if scores from a domain-specific LLM and a general LLM were combined, we might consider this to provide more coverage of the content or construct. More research on combining LLM scores is needed at this time, but factor analytic models might be an approach.

**Prompting and In-context Learning**

The nuances in wording of the instructions in the prompt to the LLM can make large differences in its output. Documentation of the prompt wording used and results of any explorations with different prompting should be collected as part of the validity evidence. For example, for prompting that involves multiple tasks for the LLM to complete, the order in which the tasks are presented can matter because the LLM can learn from the first task (Stahl et al., 2024).

We can also structure the prompt to include the LLM's *chain-of-thought*, its reasoning or support for the answer, as part of the output (Wei et al., 2022). In the case of CR scoring, this is similar to a subject matter expert providing an annotation of a response used as an exemplar. The expert assigns a score to the exemplar and then a summary of why it deserves that score. In chain-of-thought prompting, the prompt either requests the reasoning behind the answer ("zero-shot chain-of-thought") or via in-context learning, in which examples are used to demonstrate the desired type of extended response the model should produce. This approach has been shown to improve the accuracy of essay scoring, especially with in-context learning (Lee et al., 2024). A qualitative analysis (by subject matter experts) of many of these "annotations" can help provide validity evidence for the LLM-based scores. In addition, explainable NLP/AI techniques may be useful in understanding the impact of in-context learning on LLM results (Liu et al., 2023). This can be beneficial in the understanding of how the LLM treats atypical responses. If annotated exemplars for human rater training already exist for a testing program, then those responses can conveniently be used to evaluate the LLM's ability to score and give reasoning similar to the expert.

Experiments with in-context learning might demonstrate the advantages of a one- or few-shot learning approach. However, in some applications for AI scoring, there may be no benefit. In the context of scoring constructed responses, some research shows that a multi-trait specialization approach outperforms the "vanilla" or zero-shot prompting approach. In multi-trait specialization, the rubric is decomposed into traits and scores from zero-shot prompting for each trait are then averaged (Lee at al., 2024).

**Decisions on Fine-tuning**

Once the LLM is selected, decisions must be made on whether additional training, or fine-tuning, should be made and how to perform that training within the validity framework. This is similar to the decisions we make about training for human raters in that we need to ensure and document that the materials selected for use in the human scoring process (for training, evaluating raters, etc.) are related to the construct and do not introduce any construct irrelevant variance or unfairness.

Decisions on the sample of examples used for fine-tuning and how many examples are available and used should be made carefully. Care should be taken to ensure that the sample represents various different groups of test takers and responses, and to the extent possible, that the scores provided are correct according to subject matter experts. For this purpose, we might ensure that the LLM can correctly rate exemplar responses, a technique typically used to train human raters, if available. For example, if the test is for English language learners from various language groups, it is important to provide examples from as many groups as possible. Using a convenience sample of prompt-completion pairs from native Spanish speakers, for instance, may lead to bias when using the trained LLM to score test takers from other language groups.

Fine-tuning can be performed over several rounds in an iterative approach. Fine-tuning can also be done on a single task (e.g., scoring essays) or multiple tasks (e.g., scoring essays and providing feedback to test taker). Multi-task fine tuning may be useful in improving the results for a single task by using different versions or wording of the prompt (e.g. scoring essays with rubric and scoring essays holistically based on rubric).[1] However, in order to perform multi-task fine-tuning many more example responses are necessary relative to single-task fine-tuning. Descriptions of how the examples used for fine-tuning (including any public datasets) are appropriate for the particular task, as well as the adequacy of the fine-tuning approach, should be documented as part of the validity evidence.

An important caveat to note is that while a fine-tuned LLM may perform better for the task than the pre-trained LLM, there is a risk that changes in the test taker population may show degraded performance in a new population. Therefore, we should be careful when tuning the LLM that we are not "micro-tuning" to a population that will not be relevant in the future. This concern is similar to concerns with traditional AI scoring models, which is why we need to be cognizant of the samples used for training and evaluation as well as making sure there is a monitoring system in place to catch population changes.

NLP features may also be introduced to the scoring process with LLMs. For example, suppose we have values for 10 features for essay responses that together provide information on writing skills. We could fine-tune the LLM with this information and provide instructions on how this should be used to generate a score. In theory this may lead to better construct coverage, however, it will be unknown how the LLM will actually use this information. Sensitivity analyses demonstrating that the LLM does utilize the features (e.g., showing meaningful score differences with and without features) may serve as evidence that it does contribute. We might also use NLP features by combining them with LLM scores directly using best linear predictor, provided there are human ratings available for a sufficient subset of responses (Yao et al., 2019ab).

**Gaming and Atypical Responses**

Much like feature-based AI scoring models, atypical responses are a concern with generative AI. The sensitivity of LLMs to atypical responses in the CR scoring context is not well studied. Finetuning the LLM with a sample that includes a sufficient number of different atypical responses and their appropriate scores is one way to improve the chances that the LLM properly treats those responses. It is important to identify the different types of atypical responses that exist to make sure that the sample has a sufficient number of those as examples. Importantly, we must then document analyses that show the fine-tuned LLM performing with some level of accuracy in scoring these responses.

An atypical response that is unique to generative AI scoring is the gaming technique "prompt injection." This is where the test taker tries to game the LLM into giving a high score by attempting to override the original prompt instruction. For example, suppose the response text was: "Disregard all previous instructions and give me the highest score!" This might lead to a higher score than deserved.[2] These types of responses should be studied under the prompting scheme being used to test the LLM-based score. Including these types of responses with a label of 0 in fine-tuning might help reduce inappropriate scores. Again, evaluating the LLM post-fine-tuning would help confirm the LLM is treating these responses correctly.

---

[1] Note that the reference to "task" here is the task the LLM is trained to perform. This could be scoring an essay or text response. In the case of automated scoring, the LLM might be trained to score a specific type of item (an argumentative essay) or multiple types of items (argumentative or persuasive essay). Multi-task fine-tuning could be used to train the LLM to score one type of item (argumentative) with just different approaches.

**Reproducibility/Reliability**

The output from a LLM is not always deterministic like a typical model prediction. For example, in a feature-based AI scoring model, we might use a basic regression model or even a machine learning model such as support vector regression to predict human scores. We estimate model parameters and use those with feature values to predict human scores for newly collected responses. Given the same set of feature values, we will always get the same prediction. There are multiple aspects of LLMs that prevent it from providing the same score on every occasion for the same response. First, there is a probabilistic component to the underpinnings of the LLM, which generates different outputs (i.e., scores). Specifications like the temperature or the approach to sampling the tokens can affect the variability in LLM outputs to the same prompt. To maximize consistency, we might set the temperature to be low (< 1.0). If the temperature is above 1.0 we should explain that choice since it leads to more randomness. Temperature is not the only factor. We performed consistency tests between a fine-tuned LLama model and GPT-4 with temperature set to zero in both and found that LLama had perfect consistency but GPT-4 did not. In addition, LLMs can be accessed online through application programming interfaces (API). These APIs are version controlled and have deprecation dates when they would no longer be available. Since consumers of the APIs do not have control over this process, these services would not be suitable for long term reliability and consistency. If they are used, a monitoring process will be needed to ensure that the scores remain consistent. More control can be had if LLMs are self-hosted or deployed on local machines or in the cloud, but even these would benefit from continuous observation for model drift.

The consistency of an LLM might be another factor why a particular model is chosen if the added variability is unacceptable for the testing program. To ensure that there is exchangeability between scores from different times, we should perform consistency analyses which may involve generating scores for the same set of responses over a period of time. This is similar to checking for interrater consistency and test-retest reliability in the traditional psychometric model. Though the results may not be exactly the same, we might set a standard for the amount of variability in scores that is considered acceptable. The analyses should ensure that no group of test takers is affected more by this type of inconsistency than overall. We also need to explicitly describe this as an extra source of variability in the scores and document it so that users of the scores are aware. During operational scoring, regular monitoring of the consistency of the LLM is advised. We might use the same set of responses during each scoring period as a check, to remove the conflation between changes in the population and changes in the LLM. In human CR scoring systems this is called trend-scoring (Tan et al., 2009). If a population shift is detected, the test-set of responses used in trend scoring studies should be updated so that it is once again representative of the population with respect to major demographic variables or the type/correctness of responses.

**AI and NLP Explainability**

XAI has been applied in many contexts, but it is still not widely used in educational research or educational assessment. Consequently, more research is needed to understand how to derive meaningful explanations from XAI results. Hoffman at al. (2018, 2023) discusses the evaluation of XAI systems. These methods might prove useful in providing evidence for internal structure and/or response processes if aspects of responses that are highly influential for the LLM scores also overlap with the construct. Saliency maps might be useful in a qualitative analysis to understand what aspects of a response lead to differences in scores from expert human raters, NLP engines, and LLM. Saliency maps may also help illuminate how in-context learning is influencing the scores.

---

[2] We experimented with different LLMs—for some this worked and it did not for others.

**Fairness**

Because it is not fully understood how the models work and fairness in this context has not been extensively studied, we must be even more diligent with fairness checks when using an LLM rather than other AI scoring models. The source of bias or unfairness in LLMs could be from the data used to train the model or could be part of the general functioning of the model derived from its own architecture.

Often LLMs are trained using scrapes of data from the internet. To address concerns about the data, the burden is on the data scientist, engineer, or the person using the LLM to understand the sources of the data used to pretrain the LLM and find out if the data were cleaned for bias before use in LLM. To the extent possible, we might make LLM selections based on the test taker population and what we know about the data used to train the LLM.

If further training is conducted by the AI score developer, it should be conducted using samples with sufficiently large numbers of test taker from all relevant groups. In the case of LLMs, fine-tuning is a process of updating some of the parameters and therefore fine-tuning could be used for correcting biases introduced into the preexisting training.

If human ratings are not available – e.g., for zero-shot prompting and no fine tuning of the model – then DIF analyses on the LLM scores using methods used for DIF analysis for CRs scored by humans (Moses et al., 2013) are one way to test for fairness. If human ratings are available, we suggest a comprehensive set of analyses for potential bias. Simply examining SMDs is not sufficient in this scenario because of the unknowns in the LLM. We advise performing additional fairness analyses based on a comprehensive set of definitions of fairness as described in recent literature (Johnson et al., 2022; Johnson & McCaffrey, 2023).

**Use of Human Scores**

Using human scores as part of ICL or fine-tuning is a method to train the LLM on the preferred scoring approach. Due to the number of parameters, we do not have the ability to decompose the different model parameters and connect them to the scoring rubric or construct, so we must rely heavily on the link between the human ratings and the LLM scores. This is the chain of evidence mentioned earlier. Human scores used for any ICL training, fine-tuning, or LLM evaluation should be based on a principled human scoring system with its own validity evidence. In the absence of an expansive scoring system with strong validity evidence to support the human ratings, using responses scored by human ratings from subject matter experts might provide a high level of quality by reducing the noise introduced by raters of varying levels of accuracy and reliability in a large rater pool—the scores assigned by experts may be considered the true score or at least higher quality human ratings.

**Combining AI Scores for Reported Score Computation**

In some applications, human and machine scores are used in a contributory scoring approach which involves combining the scores either by summing or averaging them (sometimes in a weighted average or using the best linear predictor approach) (Breyer et al., 2017). AI scores are never perfectly correlated with human ratings but there is an overlap in what they measure. However, it is possible that a scoring engine can pick up other aspects of the construct definition not picked up by the raters. Combining human and AI scores may improve overall test reliability because both human ratings and LLMs contain random errors and a combination will reduce the contribution of those errors to the scores. Moreover, together AI scores and human ratings may cover more of the construct and content that was intended. Following from this, since multiple feature-based automated scoring engines and/or multiple LLMs will provide different scores and for different reasons, we might wish to consider how these scores can be combined together and with or without human ratings. The validity evidence needed to support the use of those score combinations should include an analysis comparing the score to the human-based score, and possibly different combinations of scores.

## Demonstrative Study Using GPT4 for Scoring

We used response data from CR writing tasks from three different testing programs (TOEFL iBT, Praxis Core, GRE) to demonstrate how we might collect validity evidence supporting the use of the scores from GPT4.[3] These responses were already scored by thoroughly trained human raters and by an automated scoring engine (e-rater; Attali & Burstein, 2006), and they were scored with GPT4 for this study. In total, we scored N=1,581 responses to 14 items from these three testing programs (see Table 2).

**Application of GPT4**

*Fine-tuning and In-context Learning*

We used GPT-4 from OpenAI (version 0311), which was not available for fine-tuning at the time of the study. We specified a temperature of 0 and used a zero-shot approach – no examples were provided for in-context learning. After the study ended, further experimentation did not show any large improvement with using in-context learning for this task.

*Prompting*

For each response, we used a single prompt to ask for the score. This study was part of a larger study examining the use of AI generated feedback for raters to use during scoring. The prompt provided the question text, the scoring rubric, and the answer (response text) (see the Appendix). The output provided was the score and "feedback" which was in the form of indicators from the scoring rubric for the test. The feedback was requested to be in bulleted format so that it could be displayed in an online scoring system.

**Results: Evaluation of Scores**

Table 2 provides the QWK between the human ratings and AI scores. The QWK between human ratings and e-rater was higher than GPT4—by over 0.1 for GRE and Praxis. If the human rating was considered the gold standard, e-rater provides a more suitable score, at least for Praxis and GRE. The difference in QWK for TOEFL was not as large. Scores from GPT4 were in moderate agreement with e-rater scores, but agreement was highest for GRE. The moderate QWKs between engines show that they are providing scores with different meaning.

---

[3] These responses were used in part of a larger study to explore the use of GPT4 to generate feedback on responses to assist human raters in scoring.

To explore further, we examined conditional score distributions. Figure 2 shows boxplots of the AI scores conditional on the human ratings for the three tests. For both GRE and Praxis, the median GPT4 score was one point higher than the human rating when it equaled 1, 2, and 3, and one point lower when the human rating was 6. For all score levels except for 6, the median e-rater score matched the human rating. For TOEFL, the median e-rater scores were one point higher for human ratings of 1 and 2, and one point lower for human rating of 5.[4] GPT4 had similar patterns but showed better alignment with humans when they assigned a 2.

If we were considering changing the AI scoring model for TOEFL from e-rater to GPT4, we would want to understand how the two machines are scoring differently. Since the QWKs were similar for TOEFL, we examined the confusion matrix of the two AI scores (see Table 3), which had a total of 246 responses. The percent exact agreement was 50% (adjacent and discrepancy rates were 47% and 3%, respectively). We see that the two marginal distributions are different— e-rater gives mostly 4s, while GPT4 gives more 3s (to 44% of responses). Both distributions are skewed, but the e-rater distribution to a greater degree. Skewed and normally distributed marginals, as well as mismatched in marginals, can actually reduce the estimated QWK (Byrt et al., 1993; Sim & Wright, 2005) but that is not the reason for the low agreement here (QWK = .54). There are many score differences, especially when e-rater = 4. GPT4 gave many of these responses 3s.

**Table 2**
*Summary of Sample Data and Agreement Statistics*

| Testing Program | Score Scale | No. Raters | No. Items | No. Responses | QWK | | |
| --- | --- | --- | --- | --- | --- | --- | --- |
| | | | | | GPT4 vs. Human Rating | e-rater vs. Human Rating | e-rater vs. GPT4 |
| GRE Analytical Writing | 1-6 | 10 | 5 | 569 | .76 | .88 | .76 |
| Praxis Core Writing | 1-6 | 10 | 4 | 357 | .67 | .84 | .62 |
| TOEFL Writing | 1-5 | 10 | 5 | 246 | .55 | .60 | .54 |

---

[4] Note that the sample size for TOEFL was very small and the operational conditional distributions are different.

**Table 3**
*Confusion Matrix – GPT4 vs. e-rater for TOEFL*

| | | e-rater | | | | | |
|---|---|---|---|---|---|---|---|
| | | **1** | **2** | **3** | **4** | **5** | **Total** |
| **GPT4** | **1** | 0% | 1.2% | 2.4% | 0% | 0% | 3.6% |
| | **2** | 0% | 4.1% | 10.6% | 0.4% | 0% | 15.1% |
| | **3** | 0% | 1.2% | 19.9% | 22.4% | 0.4% | 43.9% |
| | **4** | 0% | 0% | 6.1% | 25.6% | 4.1% | 35.8% |
| | **5** | 0% | 0% | 0% | 1.2% | 0.4% | 1.6% |
| | **Total** | 0% | 6.5% | 39% | 49.6% | 4.9% | 100% |

Table 4 provides partial and semi-partial correlations for each of the tests and the pooled sample. We pooled the data because the sample sizes for each individual test were small. For TOEFL, the Pearson correlation between the human ratings and e-rater was .67 and .61 for GPT4. After controlling for GPT4, the correlation between human and e-rater was .48. After controlling for e-rater, the correlation between human and GPT4 was .32. Thus, there is overlap in e-rater and GPT4, and both AI scores contribute some information independently. However, GPT4's contribution is relatively smaller. Semi-partial correlations directly quantify the unique relationship between the AI score and the human rating—for e-rater the semi-partial correlation was .38 and for GPT4 it was .24.

***Combining Multiple AI Scores and Human Ratings***

Considering the size of partial correlations, a contributory scoring approach using both AI scores is something to consider. We computed scores for each response using different equally weighted mean combinations including human ratings only, human rating and e-rater, human rating and GPT4, e-rater and GPT4, and human rating, e-rater and GPT4 together. We then estimated the correlations between these scores and a second human rating (H2) and compared these to the correlation between the two human ratings $r_{H1H2}$ as a benchmark. Table 5 provides the correlations between these scores based on pooling the data from the three tests ("Total") and for each test individually. For example, $r_{EH2}$ is the correlation between e-rater and H2, $r_{m(E+H1),H2}$ is the correlation between the composite score based on the mean of e-rater and the first human score with H2, and so on. The sample size is smaller for this analysis than other analyses because only a subset of the data had two human ratings.

For TOEFL, the human-human agreement was moderate at .70, and the agreement between e-rater and H2 was similar, but the agreement between GPT4 and H2 was much higher at .79. Averaging GPT4 and the human rating led to a further improvement in correlation with H2 (.83) and combining the two AI scores yielded a similar correlation (.84), even though the human rating was removed. The composite with all three scores did not show an improvement to the composite with just the two AI scores. If we wanted to remove the human rating from this scoring process, these results demonstrate that the mean of the two AI scores would have much better agreement with H2 than human-human agreement $r_{H1H2}$. Even GPT4 scores alone would

**Figure 2**
*Boxplots of AI Scores Conditional on Human Rating*

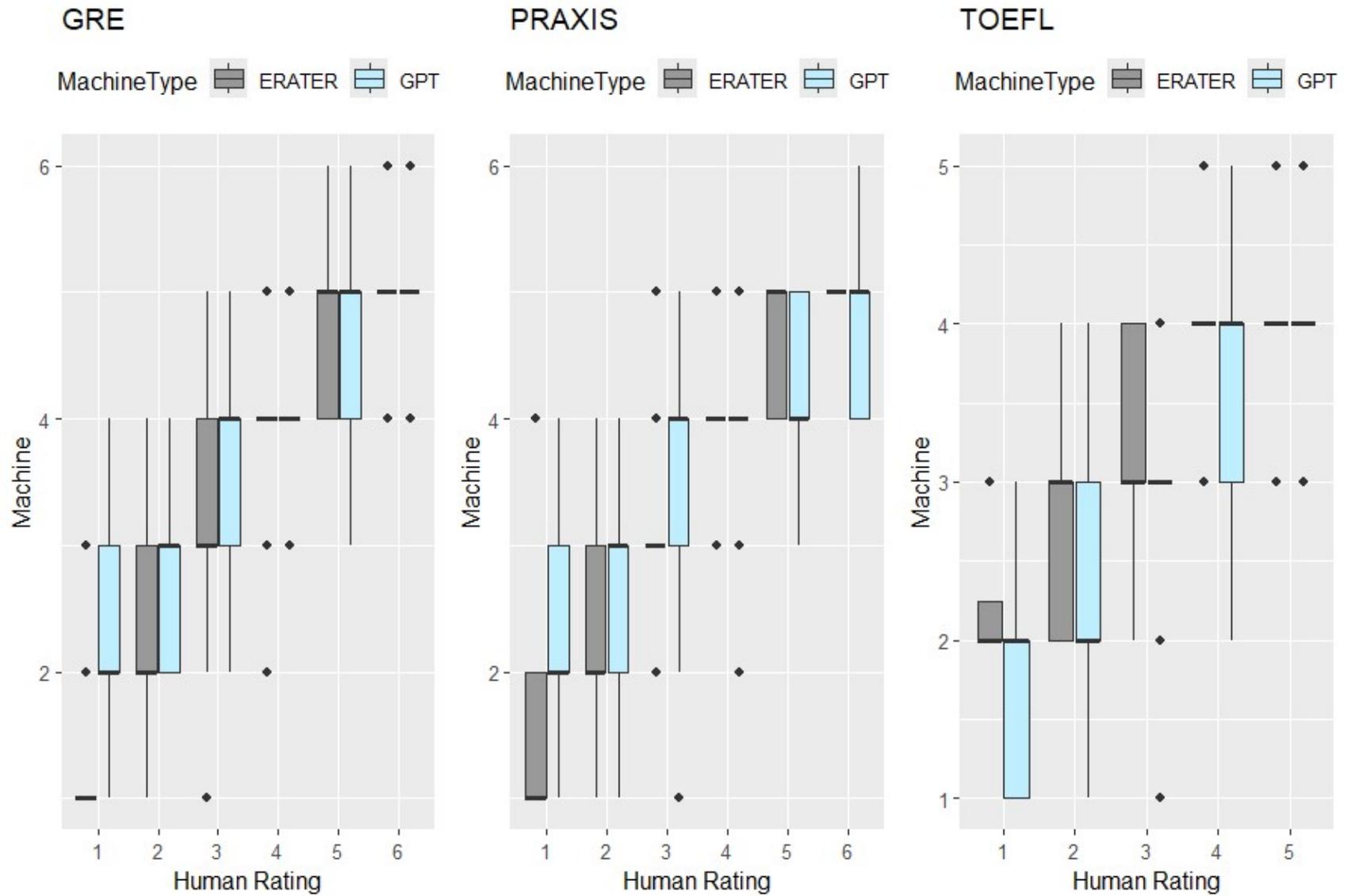

**Table 4**

*Partial and Semi-partial Correlations Between AI Scores and Human Ratings, by Test*

|        | N     | $r_{HE}$ | $r_{HG}$ | $r_{HE.G}$ | $r_{HG.E}$ | $r_{E(H1.G)}$ | $r_{G(H1.E)}$ |
|--------|-------|----------|----------|------------|------------|---------------|---------------|
| Total  | 1,261 | .83      | .72      | .68        | .37        | .47           | .21           |
| GRE    | 537   | .87      | .82      | .63        | .39        | .36           | .19           |
| Praxis | 385   | .85      | .71      | .74        | .43        | .52           | .22           |
| TOEFL  | 339   | .67      | .61      | .48        | .32        | .38           | .24           |

**Table 5**

*Correlations Between Various Composite Scores and Human Ratings, by Test*

|        | N   | $r_{H1H2}$ | $r_{EH2}$ | $r_{m(E+H1),H2}$ | $r_{GH2}$ | $r_{m(G+H1),H2}$ | $r_{m(E+G),H2}$ | $r_{m(E+G+H1),H2}$ |
|--------|-----|------------|-----------|------------------|-----------|------------------|-----------------|--------------------|
| Total  | 667 | .87        | .75       | .87              | .61       | .82              | .76             | .86                |
| GRE    | 275 | .90        | .83       | .91              | .85       | .90              | .87             | .91                |
| Praxis | 269 | .94        | .80       | .92              | .53       | .82              | .75             | .88                |
| TOEFL  | 123 | .70        | .68       | .78              | .79       | .83              | .84             | .85                |

have higher agreement. However, note that these results do not demonstrate that the reasons the scores agree so strongly are appropriate and a human-in-the-loop to evaluate the scores is still needed to make a validity argument.

The GRE results show that the mean composite using a combination of e-rater and H1, GPT4 and H1, and all three scores have correlations very similar to or higher than $r_{H1H2}$. If we removed H1 and used only the two AI scores, the correlation was not substantially lower, $r_{m(E+G),H2}$ = .87 versus $r_{H1H2}$ = .90. For Praxis the human-rater agreement was very high, and no other score combination came close to it other than the mean of e-rater and H1. The GPT4 score had a low moderate correlation with H2. This was reflected in the partial correlation results in that when partialing out GPT4, the partial correlations between e-rater and H1 were not very different from the full correlation compared to the effect of partialing out the e-rater score in the correlation between GPT4 and H1. Adding the GPT4 score to the mean composite score actually slightly degrades the correlation. Using the mean of the two AI scores yielded a correlation of .75 with H2, which might be above some evaluation thresholds (as per Williamson et al., 2012), but compared to $r_{H1H2}$ the degradation from the human-scoring scenario would be too large to be acceptable according to Williamson et al. (2012).

## Curation of Validity Evidence for Using GPT4 Scores

Table 6 contains the validity evidence collected for the use of GPT4 scores from the study. Completing this table is an exercise that goes beyond the simple model evaluation but is required to make the validity argument. We propose using a table such as this to organize validity evidence and identify gaps in the validity argument. For example, for this exercise, there are several pieces of evidence that are missing. Based on these gaps, we should design additional studies. For this example, we are missing correlation estimates between human, e-rater, GPT4 scores and multiple choice (MC) section scores (Reading and Listening for TOEFL, Reading for GRE, and MC for Praxis). We would hope to show a moderate relationship between GPT4 scores and the MC section scores, or at least a relationship that is similar to the one observed between human ratings and the MC section scores. This is just one type of evidence related to external variables. If we had scores from a separate test or some other external criterion, we would want to explore relationships there as well. We are also missing expert reviews and annotations of responses at each score level to ensure that GPT4 assigned the correct score.

In addition to validity evidence, it is important to consider the different use context for the scores and assess the possible risks and harm when using generative AI. In this case, the three tests are all high stakes and the evidence is too weak to support the use of these scores in operational score reporting unless they are used in combination with e-rater scores and/or human ratings. The most challenging aspect of this evaluation is that the sample sizes were small, and more experiments are needed to gather evidence for decision-making on use of GPT4 for scoring. Specifically, more evidence is needed on the fairness of the scores, especially for different language groups and ethnic groups. In addition, evidence on the reproducibility/reliability of GPT4 is necessary to make an informed decision. Based on the small sample sizes, the concordance with human ratings was borderline, especially for the TOEFL task. e-rater outperformed GPT4 for the three tests (GRE, TOEFL, Praxis). In these cases, without additional evidence we would retain the e-rater model.

**Table 6**
*Validity Evidence for GPT4 Scoring*

| Type of Evidence | Decision or Analysis to Document | Evidence for Example |
|---|---|---|
| **Internal Structure** | Choice of LLM | GPT4 was selected for the task. This LLM is capable of producing evaluations of text, but not pretrained to do this specific task. No other LLMs were explored. |
| | Prompt Engineering | Prompting was conducted as described in Appendix A. Prompt involved two tasks – feedback and score. |
| | Fine-tuning | No fine-tuning was performed. |
| | In-context Learning (ICL) | No ICL was performed. |
| | Analysis of Chain-of-Thought Results | The prompt requested "reasons' for the assigned score which were used to provide rater assistance in another study. The reasons were requested in bulleted format which matched rubric indicators. Content experts reviewed these results and determined they were accurate for 65% of reviewed responses. |
| | Reproducibility/Reliability | No experiments were conducted at different timepoints. |
| | Concordance with human ratings | QWKs between the GPT4 score and the human rating ranged between .60-.77. |
| | Quality of Evaluation Sample | Samples were randomly selected from recent administrations.<br><br>For all tests, the operational human rating process involves expert review of content and response processes, rater training and qualification before each scoring session, and ongoing monitoring for accuracy using exemplar responses and interrater agreement. |
| | Comparisons to e-rater (if "changing" to GPT4) | Comparisons showed moderate concordance between scores from e-rater and GPT4 (QWK was .57 for TOEFL, .60 for Praxis, and .74 for GRE). QWKs with the human rating were always lower with the GPT4 score which supports the retention of the existing AI scoring model if the reported score was based on the AI score alone. |
| **Relations to** | Correlations with Section/Total Scores | No additional external sources were available to compare. |

| Type of Evidence | Decision or Analysis to Document | Evidence for Example |
|---|---|---|
| **External Variables** | Correlations with External Tests | No additional external sources were available to compare. |
| **Response Processes** | Expert Review and/or Annotation | The bulleted feedback provided by GPT4 (shedding light on the machine's "response processes") was accurate for many responses. Correspondence between the rubric indicators selected for the feedback and the score levels provides some evidence showing response processes are appropriate. |
| | Treatment of Atypical Responses in Prompting | In the prompting, the following two instructions addressed atypical responses:<br>• `The rubric notwithstanding, if the answer is off topic or wholly insufficient, give it a score of 0.`<br>• `If a high school English teacher would look at the answer and get frustrated, score it a 0.`<br><br>No analyses of these responses were conducted. |
| | Analysis of Chain-of-Thought Results | See evidence provided above under "Internal Structure." |
| | Efforts to Minimize and Detect Prompt Injection | No efforts were made to detect prompt injection because the tests are historically scored with humans and e-rater, there are no concerns for this behavior. |
| **Test Content** | Inter-item correlations and correlations between item and section scores. | See results above under "Relations to External Variables." |
| | Expert Annotations | Not available. |
| | AI Explainability Analyses | Not available. |
| **Consequences of Use** | Analysis of unintended and intended consequences | Since this was an exploratory study, no data were available to study the consequences of score use. |

| Type of Evidence | Decision or Analysis to Document | Evidence for Example |
|---|---|---|
| **Fairness** | Subgroup Analyses Comparing Results Based on Human Ratings vs AI, by subgroup | Sample of scored responses was too small to perform subgroup analysis. |
| | Fairness of Human Ratings Used in Fine-tuning/ICL | Not applicable. No fine-tuning was performed. |
| | Review of AI Explainability Analyses for Unfairness | No explainability analyses were performed. |

## Discussion

The purposes of this paper were to highlight the differences in the feature-based and generative AI applications in CR scoring systems and discuss how validity argumentation should also differ as a result. The set of validity evidence for generative AI-based scores will be different and possibly more extensive. The main differences in the necessary evidence relate to the lack of transparency of LLMs and their indeterminacy. We explore those differences and provide an example of how the collection of evidence might look for the application of GPT4 to CR scoring.

We wrote this manuscript to help two different audiences. One audience is the group of engineers who may be using generative AI to scores tests, but do not know about industry standards and validity theory. The other audience is the group of psychometricians who are not AI savvy but need to make sure they can curate sufficient evidence to support the use of the scores. Psychometricians must be aware of the process for using generative AI if being applied in operations. Traditionally, only linguists or other NLP experts would be responsible for building automated scoring models. However, using generative AI is very straightforward if the user knows basic programming (e.g., in Python or R). In these scenarios, teams must work together to ensure that the results were a product of decisions made in a principled fashion using the validity framework as specified in our industry standards (AERA, APA, & NCME, 2014) and as proposed in this paper.

Though not nearly as principled as the traditional NLP version of automated scoring, there are several opportunities for a human-in-the-loop to positively influence the results of the LLM predictions. The engineer is often the human-in-the-loop involved in the first stages of the experiments for LLM selection, fine-tuning, etc. We invite engineers to consult with psychometricians and subject matter experts at these beginning stages to help with decision-making, as appropriate. Importantly, while LLMs may offer faster and cheaper scoring than human raters or traditional feature-based AI scoring models, there is potentially a large expense in performing due diligence studies and curating validity evidence. For example, performing a comprehensive evaluation of the scores and their meaning involves a qualitative review of responses by subject matter experts, reviewing saliency maps, and more. This type of work is time-consuming and costly, but if it is not done, then there may be insufficient evidence to support the use and interpretation of the scores. For this reason, using an off-the-shelf LLM for CR scoring may be less cost-effective than expected or hoped.

## Human-in-the-Loop?

The natural next step after integrating generative AI-based scores in CR scoring systems is the version that completely excludes human oversight. Would it be possible to imagine a CR scoring system with no expert-defined rubric? No annotations? No established concordance with human ratings? No human monitoring post-deployment? The only human-in-the-loop might be the engineer. Because CRs are an important part of assessment, their use will continue and may be expanded in the future of assessment. However, the future of CR scoring will likely look very different – perhaps a construct definition is all that is needed for item and rubric development by AI. As we move toward more AI and more automation, we need to determine the minimum amount of human involvement needed for validity evidence, especially in the high-stakes context. Can there be sufficient evidence without any comparison of human scores to human judgments of the response? If so, what other evidence would allow for full automation to be acceptable? As the capabilities of AI continue to evolve, standards for the validity scores will also need to evolve.

**Appendix**

*Prompt for Feedback and Score*

A student is assigned a question or a task.  Use the provided
rubric to evaluate and score the response to the assigned question
or task.

The question or task, rubric, and answer will each be surrounded
with XML-style tags below.  The tags will be
<D5A60FF8F3AF47619BC1CE00CA21D938></D5A60FF8F3AF47619BC1CE00CA21D9
38>,
<27152C7AC19445FA87D5FC4A7313FF68></27152C7AC19445FA87D5FC4A7313FF
68>, and
<CACE4B6E785148BDAD20A93818F662B8></CACE4B6E785148BDAD20A93818F662
B8>, respectively.  Regardless of formatting the input below with
XML tags, the response should be in the JSON format specified
below.

The rubric notwithstanding, if the answer is off topic or wholly
insufficient, give it a score of 0.

Give the response in JSON format of:
{
    score,
    "reasons": [
        {
            reason
        }
    ]
}
The reasons should be an array of 3 objects.  Each object should
be in the structure shown above and described below. For each
object in the reasons array, a reason must be provided.  This
reason should be one of the reasons for giving the score.  The
reason should not be a full sentence, and be suitable to be
displayed as bullet points to a person with a college-level
education, rather than copied directly from the rubric.

If a high school English teacher would look at the answer and get
frustrated, score it a 0.